\documentclass{new_tlp}
\usepackage{mathptmx}

\newcommand\bcmdtab{\noindent\bgroup\tabcolsep=0pt%
  \begin{tabular}{@{}p{10pc}@{}p{20pc}@{}}}
\newcommand\ecmdtab{\end{tabular}\egroup}

  \title[Management of caching for stream reasoning]
        {Managing caching strategies for stream reasoning with reinforcement learning}

        \author[C. Dodaro, T. Eiter, P. Ogris, K. Schekotihin]
        {
          CARMINE DODARO\textsuperscript{1}, 
          THOMAS EITER\textsuperscript{2}, PAUL OGRIS\textsuperscript{3}, and
          KONSTANTIN SCHEKOTIHIN\textsuperscript{3}\\
          \textsuperscript{1} Department of Mathematics and Computer Science, University of Calabria, Italy\\
          \email{dodaro@mat.unical.it}\\
          \textsuperscript{2} Institute of Logic and Computation, KBS
          Group, Vienna University of Technology, Austria,\\
          \email{eiter@kr.tuwien.ac.at}\\
          \textsuperscript{3} Alpen-Adria-Universit\"at, Klagenfurt, Austria,\\
          \email{\{paul.ogris,konstantin.schekotihin\}@aau.at}
        }
\jdate{March 2003}
\pubyear{2003}
\pagerange{\pageref{firstpage}--\pageref{lastpage}}
\doi{S1471068401001193}

\usepackage{amssymb}
\usepackage{amsmath}
\usepackage{graphicx}
\graphicspath{{figures/}}

\usepackage{amsmath}
\usepackage{amstext}
\usepackage{amssymb}
\usepackage{amsfonts}
\usepackage[amsmath,thmmarks]{ntheorem}
\usepackage{listings}
\usepackage{comment}

\newtheorem{definition}{Definition}

\newtheorem{example}{Example}
\usepackage[inline]{enumitem}
\usepackage{wrapfig}

\usepackage[ruled,vlined,linesnumbered]{algorithm2e}
\usepackage{color}
\usepackage{url}

\usepackage{todonotes}

\newif\ifdotikz\dotikztrue

\ifdotikz
\usepackage{pgf}
\usepackage{tikz}
\usetikzlibrary{arrows,positioning,automata,decorations,fit,backgrounds}
\usepgflibrary{shapes.geometric} 
\usetikzlibrary{shapes.geometric} 
\usepgflibrary{decorations.pathmorphing} 
\usetikzlibrary{decorations.pathmorphing} 
\usepgflibrary{decorations.text} 
\usetikzlibrary{decorations.text} 
\usetikzlibrary{matrix}
\pgfrealjobname{main} 
\else
\ifpdf
\long\def\beginpgfgraphicnamed#1#2\endpgfgraphicnamed{\includegraphics{#1}}
\else
\usepackage{epsfig}
\long\def\beginpgfgraphicnamed#1#2\endpgfgraphicnamed{\epsfig{file=#1.eps}}
\fi

\newcommand{\medblacksquare}{\mathbin{\scalebox{0.6}{\ensuremath{\blacksquare}}}} 

\makeatletter%
\@ifclassloaded{IEEEtran}%
{
	\newtheorem{definition}{Definition}
	\newtheorem{example}{Example}
}%
{}%
\makeatother%

\renewtheorem{definition}{Definition}
\theorembodyfont{\normalfont}

\theoremsymbol{\ensuremath{\medblacksquare}}
\renewtheorem{example}[examplecounter]{Example}
\theorembodyfont{\normalfont}
\theoremheaderfont{\normalfont\itshape}
\theoremseparator{.~}
\theoremsymbol{\ensuremath{\Box}}

\newcounter{myenumctr}

\newenvironment{myenumerate}{\begin{list}{ {\bf(\arabic{myenumctr})}\ }{\usecounter{myenumctr}
\setlength{\topsep}{0pt}
\setlength{\leftmargin}{0pt}
\setlength{\itemsep}{0pt}
\setlength{\parsep}{0.15\baselineskip}
\setlength{\itemindent}{1.35\labelwidth}}}
{\end{list}}

\newcommand{\cA}{\ensuremath{\mathcal{A}}}

\newcommand{\cP}{\ensuremath{\mathcal{P}}}

\definecolor{dark-gray}{gray}{0.25}

\makeatletter
\lst@Key{countblanklines}{true}[t]%
{\lstKV@SetIf{#1}\lst@ifcountblanklines}

\lst@AddToHook{OnEmptyLine}{%
    \lst@ifnumberblanklines\else%
    \lst@ifcountblanklines\else%
    \advance\c@lstnumber-\@ne\relax%
    \fi%
    \fi}
\makeatother

\lstdefinelanguage{asp}{
    breakatwhitespace=true,
    morecomment=[l]{\%},
    breakatwhitespace=true,
    commentstyle=\it\color{dark-gray},
    captionpos=b, 
    numbers=left,
    numbersep=5pt,
    numberstyle=\tiny\color{dark-gray},
    numberblanklines=false,
    countblanklines=false,
    frame=bt, framexbottommargin=5pt, framextopmargin=5pt,
    aboveskip=5pt, belowskip=5pt,
    abovecaptionskip=10pt
}

\newcommand{\myParagraph}[1]{\smallskip\noindent\textit{#1}}
\DontPrintSemicolon
\SetKwInput{Input}{input}

\newcommand{\mrestart}{\textsc{Restart}\xspace}
\newcommand{\revision}[1]{#1}

\newcommand{\reffig}[1]{Fig.~\ref{#1}}

\newcommand{\wasp}{\textsc{wasp}\xspace}

\newcommand{\ticker}{\textsc{Ticker}\xspace}

\newcommand{\ucap}{\ensuremath{\mathit{uCap}}}
\newcommand{\iucap}{\ensuremath{\mathit{iuCap}}}

\newcommand{\pr}{\Pi}                        
\newcommand{\gpr}{\pr^G}                     
\newcommand{\wpr}{\pr^W}                     

\newcommand{\dnot}{\ensuremath{\raise.17ex\hbox{\ensuremath{\scriptstyle\mathtt{\sim}}}}\xspace}                   
\newcommand{\imp}{\leftarrow}                

\begin{document}

\maketitle

\begin{abstract}
Efficient decision-making over continuously changing data is essential for many application domains such as cyber-physical systems, industry digitalization, etc. Modern stream reasoning frameworks allow one to model and solve various real-world problems using incremental and continuous evaluation of programs as new data arrives in the stream. Applied techniques use, e.g., Datalog-like materialization or truth maintenance algorithms to avoid costly re-computations, thus ensuring low latency and high throughput of a stream reasoner. However, the expressiveness of existing approaches is quite limited and, e.g., they cannot be used to encode problems with constraints, which often appear in practice.
In this paper, we suggest a novel approach that uses the Conflict-Driven Constraint Learning (CDCL) to efficiently update legacy solutions by using intelligent management of learned constraints. In particular, we study the applicability of reinforcement learning to continuously assess the utility of learned constraints computed in previous invocations of the solving algorithm for the current one.
Evaluations conducted on real-world reconfiguration problems show that providing a CDCL algorithm with relevant learned constraints from previous iterations results in significant performance improvements of the algorithm in stream reasoning scenarios.

Under consideration for acceptance in TPLP.  
\end{abstract}

  \begin{keywords}
    Stream reasoning, ASP, Reinforcement learning, Caching strategies
  \end{keywords}

\section{Introduction}

Stream reasoning is an emerging branch of AI connecting distributed
systems, databases, machine learning, and knowledge representation and reasoning (KRR) to create complex decision-making frameworks that operate on continuously changing data. 
The recently proposed LARS (Logic-based Analytic Reasoning over
Streams) framework  \cite{DBLP:journals/ai/BeckDE18} is rooted in
logic programming, with reasoners that allow one to efficiently model and solve real-world problems.

To ensure high throughput and low latency of a decision-making system, stream reasoners use 
various techniques to reduce the impact of redundant computations while updating their internal state as new data arrives. 
\textsc{Laser} \cite{DBLP:conf/semweb/BazoobandiBU17} uses a Datalog-like fixed-point materialization of restricted (plain) LARS formulas combined
with specific annotations of rules to avoid unnecessary
re-evaluation for subsequent portions of the data stream. Although
\textsc{Laser} demonstrates very high performance in the evaluations,
the expressiveness of the plain LARS programs handled, which is
restricted to stratified negation, is rather limited. A larger
fragment of plain LARS, which enables modeling of problems with
multiple models but no constraints, is supported by
\textsc{Ticker} \cite{DBLP:journals/tplp/BeckEB17}. 
This stream reasoner uses a justification-based truth maintenance method
\cite{DBLP:journals/ai/Doyle79} to first
retract parts of previously computed decisions that are
inconsistent with new data in the input stream and extend then the partial assignment obtained to a model.
To provide full support of plain LARS without sacrificing much latency
or throughput, \citeN{DBLP:journals/tplp/EiterOS19} presented a
distributed processing approach. In particular, their distributed reasoner uses
stream stratification \cite{DBLP:journals/ai/BeckDE18} to decompose
LARS programs into
subprograms that can be evaluated by different instances of an 
ASP solver in parallel. 

On the one hand, applying full-fledged ASP solvers in stream reasoning settings
provides an easy way to model and solve various problems
in practice, but on the other hand poses a big challenge to their reasoning algorithms. 
The Conflict-Driven Constraint Learning (CDCL)-based algorithms
used in Boolean satisfiability (SAT) \cite{DBLP:conf/ictai/SilvaS96} 
and ASP \cite{DBLP:conf/lpnmr/AlvianoDFLR13,DBLP:journals/aim/KaufmannLPS16}
are not suitable for continuous operation over a long time. 
By design, they have no specific means to efficiently manage their
internal state while solving sequences of instances 
appearing from an input stream. 
Instead, they are geared to find one or multiple solutions of a single problem instance per run, regardless of whether one-shot or incremental/multi-shot
\cite{DBLP:conf/sat/NadelR12,DBLP:conf/lpnmr/AlvianoDLR15,DBLP:journals/ijait/AudemardS18,DBLP:journals/tplp/GebserKKS19}
solving is used. 
In the latter mode, solvers incrementally construct a model for a sequence of instances, where the next one is added to those previously stored.
Thus, if any of the accumulated instances is unsatisfiable, 
the unsatisfiability will prevail. To avoid this,
modern solvers like \textsc{oclingo} \cite{DBLP:conf/kr/GebserGKOSS12} or \textsc{glucose} \cite{DBLP:journals/ijait/AudemardS18} keep all constraints that are
inconsistent with the current instance in memory, in a deactivated
state. 
Similarly, in other logic programming paradigms, such as XSB Prolog, reasoning over streams can be implemented using incremental tabling, which can efficiently track atoms appearing in the data stream and update relevant cached goals \cite{DBLP:journals/tplp/SwiftW12}.
This allows an incremental solver to keep all learned
constraints and exploit this information to solve the instance obtained in the next iteration.

The main drawback of the incremental strategy in streaming scenarios,
however, is that a solver might forget constraints learned during restarts
that are relevant for solving the next instance. 
For instance, a typical application of stream reasoners for cyber-physical systems (CPS) is to monitor and reconfigure a system such that it can react suitably to changes in its environment. 
In such a scenario, a part of the CPS may go down due to a failure or
for maintenance and then be up again after a while, i.e., the system is back to its normal state. 
Thus the reasoner must reconfigure multiple times and should not forget the constraints learned for the normal system state, as instances corresponding to it occur most often in the data stream. 

To address these issues, we make  in this paper the following contributions:
\begin{myenumerate}
  \item We present a reinforcement learning approach aiming at the identification of learned constraints having the highest utility for the overall stream reasoning process. 
  Depending on the value of the learning rate parameter, the learner can make different assumptions about the utility of learned constraints.
  Thus, given a learning rate close to 1, the learner assumes that data appearing in the stream at some point in time is closely related to the data appearing in the next time point. That is, any subsequent state of a CPS is highly related to a previous one. Therefore, data cached while solving a problem instance can help to solve the subsequent one. 
  In turn, given a relatively small learning rate, the learner gets more skeptical and modifies its estimates very slowly. In the CPS case, this means, for instance, that the learner expects the system always to return to one of its most frequent operation states after all issues that occurred have been resolved.
  \item We present a method that can efficiently cache and manage data computed by a solving algorithm. 
    For our reference implementation, we extended the \textsc{wasp} solver
    with functionality for data exchange with external caches and equipped it with an overgrounding approach as in \cite{DBLP:journals/tplp/CalimeriIPPZ19}.
  \item We conduct an extensive evaluation and parameter tuning of the suggested approach using a version of the Partner Unit Problem (PUP)
    \cite{DBLP:conf/cpaior/AschingerDFGJRT11}, which represents a
    continuous operation (reconfiguration) of various safety systems, and n-Queens Completion \cite{DBLP:journals/jair/GentJN17}. The results show
    that the new approach significantly outperforms existing systems
    for plain LARS using ASP solvers, like
    \cite{DBLP:journals/tplp/EiterOS19}. Furthermore, they underline
    interesting links between the working of stream reasoners and ASP
    solvers considered, and may guide the development of future systems.
\end{myenumerate}

\section{Preliminaries}\label{sec:prelim}

LARS extends ASP with specific features for various stream reasoning problems
\cite{DBLP:journals/ai/BeckDE18}. 
For instance, using \emph{window functions}\/ one can access parts of a stream such as all
data that appeared in a given time interval or the last $n$ tuples of the
stream. 
Besides windows, LARS has \emph{temporal modalities}:
\begin{enumerate*}[label=\emph{(\roman*)}]
  \item the \emph{at} operator $@_t$ where $t$ is a time point, 
  \item the \emph{everywhere} operator $\Box$, and 
  \item the \emph{somewhere} operator $\Diamond$.  
\end{enumerate*}
Plain LARS programs were translated into ASP programs
either natively using a ticked encoding  
\cite{DBLP:journals/ai/BeckDE18},
external predicates \cite{DBLP:conf/icc/BeckBDEHS17}, or functions
\cite{DBLP:journals/tplp/EiterOS19}. 
We thus consider in the sequel only techniques aiming at performance improvements of a continuously
running ASP solver in stream reasoning applications.

\myParagraph{Syntax.} 
A normal ASP program $\pr$ is a finite set of rules of the form
\begin{equation}
	a \imp l_1, \ldots, l_n
	\label{eq:rule}
\end{equation}
where $a$ is an atom (which may be absent) and $l_1, \ldots, l_n$ are literals for $n \geq 0$. An \emph{atom} is an expression of the form $p(t_1, \ldots, t_k)$, where $p$ is a predicate symbol and $t_1, \ldots, t_k$ are \emph{terms}, i.e., either a \emph{variable} or a \emph{constant}.
A \emph{literal} $l$ is either an atom $a_i$ (positive) or its
negation $\dnot a_i$ (negative), where $\dnot$ is \emph{negation as
failure}; the complement (opposite) of $l$ is denoted by   
$\overline{l}$, and we let  $\overline{L} = \{ \overline{l} \mid l \in
L\}$.
An atom, a literal, or a rule is \emph{ground}, if no variables appear
in it. The grounding of a program $\Pi$
is the set $\gpr$ of all ground rules
constructible from rules $r \in \pr$ by substituting each variable in
$r$ with some constant appearing in $\pr$.

Given a rule $r$ of the form \eqref{eq:rule}, the set $H(r) = \{a \}$
is the \emph{head} and the set $B(r) =
B^+(r) \cup B^-(r) = \{ l_1, \ldots, l_n \}$ 
is the \emph{body} of $r$, where $B^+(r)$ and $B^-(r)$ contain the
positive and negative body literals, respectively. A rule $r$ is a \emph{fact} if $B(r) = \emptyset$ and a \emph{constraint} if $H(r) = \emptyset$.

\myParagraph{Semantics.}
The semantics of an ASP program $\pr$ is given for its ground instantiation $\gpr$.
Let $\cA$ be the set of all ground literals occurring in $\gpr$. An
\emph{interpretation} is a set $I\subseteq \cA \cup \overline{\cA}$ of
literals that is \emph{consistent}, i.e., $I \cap \overline{I}=\emptyset$;
each literal $l \in I$ is true, each literal $l \in
\overline{I}$ is false, and any other literal is undefined.
An interpretation $I$ is \emph{total}, if $\cA\subseteq I\cup \overline{I}$.
%
An interpretation $I$ \emph{satisfies} a rule $r \in \gpr$, if $H(r)
\subseteq I$ whenever $B(r) \subseteq I$. A \emph{model} of $\gpr$ is
a total interpretation $I$ satisfying each $r\in \gpr$;
moreover, $I$ is \emph{stable} (an \emph{answer set}), if
$I$ is a $\subseteq$-minimal model of the reduct $\{H(r) \imp B^+(r) \mid r \in \gpr, B^-(r) \cap \overline{I} =
\emptyset\}$ \cite{DBLP:conf/iclp/GelfondL88}.
Any answer set of $\gpr$ is also an answer set of $\pr$.
By $AS(\pr)$ we denote the set of answer sets of $\pr$, which are those of $\gpr$.

\begin{algorithm}[t] 
	\SetKwInOut{Input}{Input}
	\SetKwInOut{Output}{Output}
	\Input{a ground program $\gpr$, a set of assumptions $A$, and a set of constraints $C$}
	\Output{a tuple $(I,C)$, where $I$ is an answer set or \emph{incoherent}, and $C$ is a set of constraints}
	\SetKwFunction{Propagate}{Propagate}
	\SetKwFunction{Learn}{LearnConstraints}
	\SetKwFunction{ChooseL}{ChooseUndefinedLiteral}
	\SetKwFunction{CreateC}{CreateConstraint}
	\SetKwFunction{RestoreConsistency}{RestoreConsistency}	
	\SetKwFunction{Restart}{RestartIfNeeded}
	\SetKwFunction{Delete}{DeleteConstraints}		
	\SetKwFunction{Incoherent}{Incoherent}
	$I \gets \emptyset \qquad \wpr \gets \gpr \cup \{\imp \overline{p} \mid p \in A\}$\;
	$I \gets $ \Propagate{$\wpr \cup C$, $I$}\; \label{ln:alg:propagate}
	\lIf{$I$ is consistent \textbf{and} total}{\Return $(I,\ C)$\label{ln:alg:total}}
	\uIf{$I$ is consistent}{
		$I \gets $ \Restart{$I$}\;\label{ln:alg:restart}
		$C \gets $ \Delete{$C$}\; \label{ln:alg:delete}
		$I \gets $ \ChooseL{$I$}\; \label{ln:alg:choice}
	}
	\Else{
		$r \gets$ \CreateC{$\wpr \cup C$, $I$}\; \label{ln:alg:learning}
		$I \gets $ \RestoreConsistency{$\wpr \cup C$, $I$}\; \label{ln:alg:restore}
		$C$ $\gets$ $C \cup \{r\}$\; \label{ln:alg:addconstraint}
		\lIf{$I$ is inconsistent}{\Return($\mathit{incoherent}$, $C$)}			
	}
	\textbf{goto}~\ref{ln:alg:propagate}\;
	\caption{FindAnswerSet}\label{algo:cdcl}
\end{algorithm}
\myParagraph{Conflict-Driven Constraint Learning (CDCL).} Modern ASP
solvers compute answer sets using a CDCL-based
algorithm~\cite{DBLP:journals/aim/KaufmannLPS16}
as illustrated by Algorithm \ref{algo:cdcl}.
The algorithm takes as input a ground program $\gpr$, a set of assumptions literals $A$ and a set of constraints $C$.
The idea of the algorithm is to iteratively build an answer set $I \supseteq A$ of the program
$\gpr \cup C$ resp.\ to prove that no such an answer set exists.
To this end, $I$ is initially set to $\emptyset$, and a working
program $\Pi^W$ initialized to $\gpr \cup C_A$, where $C_A$ are constraints enforcing the truth of literals in $A$.
Function \texttt{Propagate} (line~\ref{ln:alg:propagate}) extends $I$
with all literals that can be deterministically inferred. After
propagation, three cases may occur:

\begin{list}{}{
\setlength{\topsep}{1pt}
\setlength{\leftmargin}{3pt}
\setlength{\itemsep}{1pt}
\setlength{\itemindent}{5pt}}

\item[]$(i)$ $I$ is consistent and total. Then $I$ and $C$ are returned.
\item[]$(ii)$ $I$ is consistent but not total. 
	Then the algorithm uses a heuristic strategy to decide whether
        the computation must restart from scratch, to explore
        different branches of the search tree (line
        \ref{ln:alg:restart}), and whether to delete some constraints in $C$ (line \ref{ln:alg:delete}).
 \texttt{ChooseUndefinedLiteral}\/ extends then $I$ with an undefined
        literal $\ell$ (called branching literal,
        line~\ref{ln:alg:choice}) selects by some heuristics.
        A subsequent propagation step infers then the consequences of this choice.

\item[]$(iii)$ $I$ is inconsistent. Thus there is a conflict and $I$ is analyzed. 	
The reason for the conflict is modeled by a fresh constraint $r$
computed by \texttt{CreateConstraint} function (line~\ref{ln:alg:learning}).
Then the algorithm backtracks (i.e.\ choices and their consequences
are undone) until the consistency of $I$ is restored
(line~\ref{ln:alg:restore}, often called \textit{backjumping}) and $r$ is added to $C$ (line~\ref{ln:alg:addconstraint}).
If the conflict is unavoidable, i.e., the consistency of the interpretation 
cannot be restored, the algorithm terminates returning ($\mathit{incoherent}$, $C$).	
\vspace{2pt}
\end{list}
Function \texttt{CreateConstraint} is crucial for the good performance
of the algorithm. Indeed, it acquires information from conflicts and
computes a constraint to avoid exploring the same search branch
repeatedly.
However, the number of constraints added might be exponential in the
program size. Therefore, some of the learned constraints must be
periodically deleted by the function \texttt{DeleteConstraints}.
An important note is that the algorithm internally associates each literal $\ell \in I$ with a {\em decision level}, denoted $\mathit{dl}(\ell)$ and computed as follows.
Let $\mathit{maxdl}(I)=0$ if $I = \emptyset$, and $\mathit{maxdl}(I) = max(\{\mathit{dl}(\ell) \mid \ell \in I\})$ otherwise.
Then, $\mathit{dl}(\ell) = 1+\mathit{maxdl}(I)$ if $\ell$ is a
branching literal, and $\mathit{dl}(\ell) = \mathit{maxdl}(I)$ otherwise.
Decision levels are used by \texttt{CreateConstraint}.
In fact, whenever a new constraint $r$ is learned,
a positive value called \emph{Literals Blocks Distance (LBD)}\/
\cite{DBLP:conf/ijcai/AudemardS09} is associated with it 
representing the number of different decision levels appearing in $r$.
The function \texttt{DeleteConstraints}
removes constraints with large LBD values since learned constraints 
with small LBD are viewed as important.

\section{Management of caching strategies}
\label{sec:method}

Stream reasoning paradigms based on logic programming, such as LARS,
convert all incoming data into atoms (literals) and forward them to
reasoners. A single stream reasoner thus gets a sequence of atom sets 
appearing in a data stream over time. 
We assume that these sets are sampled from the same distribution, but the parameters of this distribution are unknown.

\begin{example}
    \label{ex:fault_distr}
    In many technical systems the distribution of events quite often
    follows some kind of a power-law distribution, like the Pareto
    principle -- 20\% of components of cars fail in 80\% of all
    cases. Thus, it was observed that e.g.\ faults in software
    \cite{DBLP:journals/ibmrd/Adams84} or content requested by users
    in content-centric networks \cite{DBLP:conf/infocom/RossiR12} obey
    Zipf's law.
\end{example}

Finding solutions for complex problem instances might take considerable time. Therefore, ASP solvers apply various caching techniques designed to identify, store, and reuse different results obtained by its search algorithm. 
Most of the modern solvers apply two caching techniques: constraints learning and progress (phase) saving \cite{DBLP:conf/sat/PipatsrisawatD07}.

\subsection{Constraint learning}

As discussed in the previous section, the solver analyses every
conflict found to learn a constraint preventing its reoccurrence in
subsequent steps.
During solving, the algorithm might learn many constraints and, as previous
experiments show, their number might grow very fast and thus negatively impact its performance \cite{DBLP:conf/aaai/GomesSK98,DBLP:conf/ijcai/Huang07,DBLP:journals/ijait/AudemardS18}. 
Modern solvers, therefore, adopt various restart strategies that drop unimportant constraints. 

Similarly to standard solving algorithms, the preservation of learned
constraints might improve the performance of stream reasoners, as they provide valuable information about conflicts found during previous calls, as it happens e.g.\ in Assumption-based Truth Maintenance Systems (ATMS) \cite{DBLP:journals/ai/Kleer86}.
The latter also record constraints by analyzing conflicts found during the search.
The constraints are stored in a specific database and help the reasoner to determine whether a set of new assumptions or assignments contains a known contradiction.
As a result, ATMS can significantly speed up repeated reasoning tasks. 
The main problem of ATMS is that it stores all constraints and can only drop those subsumed by recent constraints.
Modern incremental solvers instead freeze constraints unused by the solver and reactivate them when needed \cite{DBLP:journals/ijait/AudemardS18}. 
The decision -- if a constraint must be frozen or activated -- is usually made by a heuristic. 
\citeANP{DBLP:journals/ijait/AudemardS18} used a progress saving measure defined by $|\cP \cap r|$, where $r \in C$ is a learned constraint and $\cP$ is a set of literals stored by progress saving, as discussed in the next section. 

Modern stream reasoners use learned constraints only 
for one reasoning cycle, i.e.\ call of Alg.~\ref{algo:cdcl}. 
For instance, \ticker \cite{DBLP:conf/icc/BeckBDEHS17} and the
distributed reasoner \cite{DBLP:journals/tplp/EiterOS19} apply ASP
solvers to find answer streams for new incoming data. 
This approach, which we call \mrestart, creates a new instance of an ASP solver 
each time reasoning is invoked. 
Specifically, it rewrites a given LARS program $P$ into an ASP program $\pr$.
When new data appears in the input stream at time $t$, the reasoning process registers a set $L=L^{+}_t \cup L^{-}_t$ of ground atoms, where $L^{+}_t$ and $L^{-}_t$ comprise atoms that
appeared in resp.\ disappeared from the stream.
The set $L$ is used to extend $\pr$ with facts and obtain a ground program $\gpr$. 
Finally, Alg.~\ref{algo:cdcl} is run to find answer sets of $\gpr$ which correspond to the answer stream of the

A stream reasoner can store the constraints learned now and reuse them later. 
However, this might lead to 
increased memory consumption and decrease the propagation performance. 
Applying techniques from incremental solving, such as freezing/reactivating constraints directly, can be problematic. 
Their heuristics are geared to incremental answer set finding for one program, 
which may result from multiple grounding steps.
In stream reasoning, Alg.~\ref{algo:cdcl} aims to find answer sets of different
but possibly very similar ground programs for sets of atoms (dis)appearing in the input stream.

\vspace{5pt}
\noindent{\bf Heuristics.}\
Reinforcement learning (RL) can be applied to finding required heuristics using various methods \cite{Sutton2018}, which, in general, can be split into \emph{model-based} or \emph{model-free} methods. 
The former methods assume that the learning agent has multiple states and transition probabilities between these states as reactions on the actions of a learner are known. 
In the case of the stream reasoning, the states might correspond to sets of learned constraints active in the solver and transition probabilities to the likelihood that the current set will be replaced by another one when new data will appear in the stream. 
The model-free approaches do not make such assumptions. 
In this work, we focus on the latter since the development of models is quite complicated and often cannot be done automatically. 
Next, the learning methods are differentiated wrt.\ rewards. The \emph{immediate reward} methods assume that a learner gets rewards after each action, whereas in the case of \emph{delayed rewards} the learner gets feedback describing its success after a sequence of actions.
We assume that immediate rewards are more suitable for stream reasoning because the utility of a previously learned constraint for the current call of Alg.~\ref{algo:cdcl} is available as soon as it terminates.

The most known learning problem with immediate rewards is the \emph{multi-armed bandit} \cite{Sutton2018}. 
In this problem, a learner has to select one out of $k$ available actions aiming to 
maximize the expected total reward over some time period. 
The reward is sampled from some unknown probability distribution that depends on the chosen action. 
However, in our case the learner should select a subset of actions, where each action represents a learned constraint that must be unfrozen in the solver.
Therefore, we are focusing on a \emph{multi-armed bandit problem with multiple
plays} \cite{anantharamAsymptoticallyEfficientAllocation1987}, 
which can be formulated as:
given a set $N = \{ n_1,\ldots,n_k\}$ of random variables with unknown
means $\Theta=\{\theta_i=\mathbb{E}[n_i] \mid n_i \in N\}$ that are
i.i.d.\ over time, at each time point $t$ a set $N_t \subseteq
N$ is selected according to weights $W_{t-1}$ associated with 
the variables in $N$. 
The selected variables $N_t$ are observed at $t$ and a reward vector
$R_t$ is determined for them, which helps to compute a new weight set $W_t$ 
that better approximates $\Theta$.

In the context of stream reasoning, the random variables $N$
correspond to the set $C$ of learned constraints in Alg.~\ref{algo:cdcl}.  
appearing in the input stream is unknown to the reasoner, but we
stationary. 
Any atom set $L_t$ generated by 
Alg.~\ref{algo:cdcl} processes $\gpr_t$ and returns a pair $(I_t,C_t)$, 
where $C_t$ is a set of learned constraints. 
Every constraint $c \in C_t$ is associated with a reward $R_t(c)$, 
which depends on whether Alg.~\ref{algo:cdcl} used $c$ for computing
answer sets (positive) or not (negative reward). 
The learning algorithm uses these rewards to update its estimate $w_t^c$ of the expected reward for unfreezing $c$ in the solver.
The goals is to find a set $W$ of estimates of expected rewards for all known learned constraints, called
\emph{policy}, that maximizes the (weighted) sum of
rewards at all time points when Alg.~\ref{algo:cdcl} is
run; i.e., a policy should maximize the probability to select
and activate a subset of constraints learned at times
$1,\dots,t$ for propagation while finding answer sets

Similarly to \citeN{gaiCombinatorialNetworkOptimization2012a}, we use
an action-value method that for each constraint $c \in C$ determines
its weight $w_t$ at a time point $t$ 
with an update rule: 

\begin{equation*}
w_{t+1}^c = w_{t}^c + \lambda\cdot[R_{t}(c) - w_{t}^c]
\end{equation*}
\noindent where $R_t(c)$ is a reward 
from activating/freezing 
constraint $c$ at the time
$t$ and $0<\lambda\leq 1$ is a constant determining the learning rate.
For a constant learning rate, the update rule can be reformulated in a non-recursive form:
\begin{align*}
    w^c_{t+1} & = 
(1-\lambda) w^c_t + \lambda R_t(c) =
(1-\lambda) ((1-\lambda) w^c_{t-1} + \lambda R_{t-1}(c)) + \lambda R_t(c) = \\
& = (1-\lambda)^2 w^c_{t-1} + (1-\lambda) \lambda R_{t-1}(c) + \lambda R_t(c) = \dots
= (1-\lambda)^t w^c_1 + \lambda \sum_{i=1}^t (1-\lambda)^{t-i} R_i.
\end{align*}
Consequently, the learning method focuses on the latter rewards and gives increasingly higher discounts for old rewards. 
As a result, the longer a learned constraint is not used by the solver, the higher is the likelihood that the learner will advise the solver to delete it during the next restart.  

Depending on the definition of the reward function, we can obtain different estimates $W$ of the optimal policy $W^*$.
In this paper, we consider the following reward function:
\begin{equation*}
    R_t(c) = a \cdot[1 - 2 \cdot \mathit{LBD}_t(c) +
    \mathit{ua}_t(c) - \mathit{uf}_t(c) - 0.25\cdot\mathit{nf}_t(c)]
\end{equation*}
\noindent where 
\begin{enumerate*}[label=\textit{(\roman*)}]
    \item $\mathit{LBD}_t(c)$ is the value of the LBD heuristic computed by Alg.~\ref{algo:cdcl} for a learned constraint $c$;
    \item $a$ is a coefficient selected wrt.\ the number of decision levels of a ground program, with $a = 20$ in the experiments;
    \item $\mathit{uf}_t(c) = 1$ if $c$ was frozen, i.e.\ not
    initially provided to Alg.~\ref{algo:cdcl}, but rediscovered during its execution;
    \item $\mathit{ua}_t(c) = 1$ if a constraint was provided to Alg.~\ref{algo:cdcl} and used by it; and
    \item $\mathit{nf}_t(c) = 1$ if the constraint was frozen and not
    rediscovered.
\end{enumerate*}
The coefficient $\mathit{nf}$ allows a learner to penalize and subsequently remove constraints that were frozen for a long period of time.

Finally, we use the optimistic initial values \cite{DBLP:conf/nips/Sutton95} as the exploration strategy. 
That is, a learner as formulated above uses estimates of expected rewards to decide which constraints must be unfrozen in the solver before each call to Alg. \ref{algo:cdcl}. 
By specifying initial values $w_1^c$ much larger than the possible values of the reward function, we encourage the learner to use newly found constraints more often in order to determine good estimates for their expected rewards.
For instance, if a constraint is learned at a higher decision level, it might get a high LBD value in the first reward. 
As a result, this constraint will never be unfrozen again. 
Optimistic initial values of $w^c_1$ allow the learner to avoid such cases. 

\subsection{Progress saving} 

Progress saving is a caching technique that stores assignments by the
search algorithm to avoid recomputations caused by backjumping (non-chronological backtracking)
\cite{gaschnigjohnPerformanceMeasurementAnalysis1979}. 
As practice shows, the latter may cause the deletion of assignments unrelated to found conflicts. 
Thus solutions of subproblems might be computed again. 
Progress saving can avoid this by keeping an array of literals deleted during the
backjumping and using it for branching decisions.

The effect of progress saving is similar to the one of JTMS techniques \cite{DBLP:journals/ai/Doyle79} used in the \ticker reasoner \cite{DBLP:conf/ijcai/BeckDE15}. 
The latter labels each rule with a time interval in which it should be considered by the reasoner, which determines the interval by analyzing the rule body at 
whenever new data appear in the input stream.
When a rule is activated, it is materialized and moved to a cache.
At the same time, all rules with expired labels are removed.
Just as the progress saving technique, \ticker always stores the model
from the last reasoning iteration.  
At each reasoning call, \ticker updates the model by removing all literals derived from removed rules and adding new literals via propagating activated rules.
Implementations of Alg.~\ref{algo:cdcl}, e.g.\ \wasp, can store the cache of progress saving over multiple calls.  
When underlying assumptions change, \wasp can use this cache to restore the last model if the new assumptions are satisfied.

\subsection{Implementation details}

The discussed caching strategy based on reinforcement learning was implemented as shown in Alg.~\ref{algo:management}. 
Prior to executing the main algorithm, an overgrounded program $\gpr$ is generated.
Just as in \cite{DBLP:journals/tplp/CalimeriIPPZ19}, we disable all rule simplifications of a grounder and then provide it with sets of possible facts corresponding to all ground atoms possible in the input data stream.
For reconfiguration problems that we consider in this paper, the required set of possible constants is finite and it corresponds to the set of CPS components. 
The resulting ground program includes all rules that can be generated for all possible instances of the given problem encoding.
As our experience shows, a complete overgrounding performed by our method by default allows a solver to run all simplification, preprocessing, and other routines at once. This enables time savings when new data appears in the stream. The original approach of \citeN{DBLP:journals/tplp/CalimeriIPPZ19} can however be considered when a complete overgrounding is too large. Especially this approach might be interesting in applications where decisions must be made depending on values measured by CPS sensors or computed by other subsystems.

Alg.~\ref{algo:management} starts a solver and does overgrounding during initialization which may take longer than the grounding step of \mrestart.
During operation, Alg.~\ref{algo:management} identifies new
assumptions using the ground atoms from the input stream and
determines $k$ constraints to be activated in the solver.  
Next, it calls Alg.~\ref{algo:cdcl} that finds answer sets as well as
statistics required to compute the rewards. 
Finally, the weights of the learned constraints are updated and
a new portion of ground atoms is read from the stream.
In our experiments, we kept the cache size small -- $k=3000$ and $n=2 \cdot k$ -- to ensure efficient execution of the propagation in Alg.~\ref{algo:cdcl}.

\begin{algorithm}[tbh]
    \SetKwFunction{reads}{ReadStream}
    \SetKwFunction{outs}{WriteStream}
	\SetKwFunction{ProcessAssumptions}{UpdateAssumptions}
	\SetKwFunction{DescendingSort}{DescendingSort}
	\SetKwFunction{From}{From}	
	\SetKwFunction{FindAnswerSet}{FindAnswerSet}
    \SetKwFunction{ComputeReward}{ComputeReward}	\SetKwInOut{AlgoInput}{input}
    \SetKwInOut{params}{parameters}
    \SetKwInOut{AlgoOutput}{output}\SetKwInOut{AlgoPersist}{global}
    \SetKw{Break}{break}
    \SetKw{goto}{goto}
	
    \AlgoInput{a ground program $\gpr$}
    \params{learning rate $\lambda$, a number $n$ of constraints to
      store, and a number $k$ of constraints to activate}
	\AlgoOutput{an answer set $I$ or $\mathit{incoherent}$}
    \AlgoPersist{constraints set $C = \emptyset$, assumptions set $A = \emptyset$, list of weights $w = [\,]$}
    $(L^{+}, L^{-}) \gets$ \reads() \;
	$A \gets $ \ProcessAssumptions($L^{+}$, $L^{-}$, $A$)\;
	$C^w \gets $ \DescendingSort($C$, $w$)\tcp*{Order constraints in $C$ w.r.t.\ $w$}
	$C_{\mathit{use}} \gets $ \From($C^w$, 0, $k$)\tcp*{Use first $k$ constraints}
	$C_{\mathit{frozen}} \gets $ \From($C^w$, $k$, $n$)\tcp*{Freeze $n-k$ constraints}
	$C \gets C\ \setminus $ \From($C^w$, $n$, $\mid C^w\mid$)\tcp*{Remove remaining constraints}
	$(I,C) \gets $ \FindAnswerSet($\gpr$, $A$, $C_{\mathit{use}}$)\;
	$C \gets C \cup C_{\mathit{frozen}}$\;
	\outs($I$)\;
	\For{$c \in C$}{
		$r \gets$ \ComputeReward($c$)\;
		\lIf(\tcp*[f]{New constraint is learned}){$c \not \in (C_{\mathit{use}} \cup C_{\mathit{frozen}})$} {$w[c] \gets w_1 + \lambda \cdot r$} 
		\lElse(\tcp*[f]{Update constraint cumulative reward}){$w[c] \gets w[c] + \lambda\cdot(r - w[c])$}
	}	
	\goto 1
		
\caption{ProcessStream\label{algo:management}}
\end{algorithm}

\section{Evaluation}
\label{sec:eval}

The evaluation of the suggested approach for various learning rates was performed on two (re)configuration problems: the partner unit problem (PUP)
\cite{DBLP:conf/cpaior/AschingerDFGJRT11}
and n-Queens completion (QC) \cite{DBLP:journals/jair/GentJN17}. 
We selected these problems because of the following reasons. 
First, (re)configuration -- known also as self-healing or resilience -- is an important topic widely discussed in the domain of cyber-physical systems (CPS), see e.g.\ \cite{DBLP:journals/cii/HehenbergerVBET16,8606923}.
Second, both problems can easily be encoded in plain LARS but, at the moment, they can only be solved using methods like \mrestart, which translate LARS encodings into ASP.  
Finally, they are well-known in the community and were used as benchmarks in ASP Competitions.\footnote{\url{https://asparagus.cs.uni-potsdam.de/contest/} or \url{https://www.mat.unical.it/aspcomp2014/}}

\subsection{Partner Unit Problem}

PUP is an abstract configuration problem with numerous industrial applications such as railroad interlocking systems, security monitoring systems, or peer-to-peer networks \cite{DBLP:conf/cpaior/AschingerDFGJRT11}.
For instance, in security monitoring applications, the goal is to ensure that at any time only an allowed number of persons are in each \emph{security zone}.
The movements of persons between rooms is registered by \emph{sensors} that are placed on doors between the rooms as well as the entrances to the building.
To ensure the security of the building, sensor readings and zone equipment must be  connected over a network of \emph{communication units}, where each unit has an equal number of ``sensor'' and ``zone'' ports defined by \emph{Unit CAPacity} ($\ucap$).
A unit-to-unit network is established using one or more communication ports,
whose number is defined by \emph{Inter-Unit CAPacity} ($\iucap$).
To ensure near real-time communication, it is required that if a zone/sensor is connected to a unit $U$, then all sensors/zones related to it must be attached either to $U$ or to other units directly connected to $U$.

\begin{example}[PUP security application]
Consider a small PUP problem in which a building has four rooms and
two entrances, shown in \reffig{fig:pup_example}. 
It has six security zones  controlling both entrances 
and all doors. 
To guarantee the security of the building, each zone should be able to
read observations of door sensors registering movement of persons,
e.g.\ the switch zone $z_{123}$ comprises sensors $s_1,s_2\text{, and }s_3$ which control all three tracks of the switch.
Zones sensors, and zone-to-sensor relations can be represented as a bipartite graph, shown in \reffig{fig:pup_example}. 
\begin{figure}[hbt]
\definecolor{darkgreen}{RGB}{68,180,46}
\definecolor{darkgray}{RGB}{80,80,80}
\centering
\begin{minipage}[b]{0.3\textwidth}
	\begin{tikzpicture}[x=2.2em,y=2.2em,baseline=0pt,rotate=90]
	\tikzstyle{track} =   [double=black!10, double distance=0.2em, rounded corners=0.4em]
	\tikzstyle{hsensor} =  [rectangle, fill, inner sep=0, minimum width=1em, minimum height=.4em]
	\tikzstyle{vsensor} =  [rectangle, fill, inner sep=0, minimum width=.4em, minimum height=1em]
	\tikzstyle{zone} =    [black!60, dashed, semithick, rounded corners=0.2em]
	
	\draw[track] (0,-1) rectangle ++(6,3);
	\draw[track] (0,.5) -- ++(6,0);
	\draw[track] (3,-1) -- ++(0,3);

	\node[vsensor, label=right:$s_1$] at (1.5,-1) {};
	\node[hsensor, label=below:$s_2$]  at (3, -.25) {};
	\node[vsensor, label=left:$s_3$] at (1.5, .5) {};
	\node[vsensor, label=left:$s_4$] at (4.5, .5) {};
	\node[hsensor, label=below:$s_5$] at (3, 1.25) {};
	\node[hsensor, label=above:$s_6$] at (6, 1.25) {};
	
	\draw[zone] (1,-1.75) rectangle node[above=1.2em, label=right:$z_{1}$]{} ++(1,1.25);
	\draw[zone] (0.25,-.75) rectangle node[below=.25em, label=below:$z_{123}$]{} ++(2.5,1);
	\draw[zone] (3.25,-.75) rectangle node[below=.25em, label=below:$z_{24}$]{} ++(2.5,1);
	\draw[zone] (0.25,.75) rectangle node[below=.25em, label=below:$z_{35}$]{} ++(2.5,1);
	\draw[zone] (3.25,.75) rectangle node[below=.25em, label=below:$z_{456}$]{} ++(2.5,1);
	\draw[zone] (5.5,.9) rectangle node[right=1em, label=above:$z_{6}$]{} ++(1.3,.7);
	\end{tikzpicture}
\end{minipage}
\hfill
\begin{minipage}{0.3\textwidth}
	\vspace{-120pt}
	\begin{tikzpicture}[x=1.7em,y=1.7em,baseline=0pt,rotate=90]
	\tikzstyle{sensor} = [draw, circle, inner sep=0em, minimum width=1.5em]
	\tikzstyle{zone} =   [draw, rectangle, inner sep=.2em, minimum width=2em, rounded corners=0.2em]
	\tikzstyle{unit} =   [draw, rectangle, inner sep=.2em, minimum width=2em]

	\node[sensor] (s1) at (0.0,1.25) {$s_1$};
	\node[sensor] (s2) at (1.5,1.25) {$s_2$};
	\node[sensor] (s3) at (3.0,1.25) {$s_3$};
	\node[sensor] (s4) at (4.5,1.25) {$s_4$};
	\node[sensor] (s5) at (6.0,1.25) {$s_5$};
	\node[sensor] (s6) at (7.5,1.25) {$s_6$};
	
	\node[zone] (z1)   at (0.0,-1.25) {$z_{1}$};
	\node[zone] (z123) at (1.5,-1.25) {$z_{123}$};
	\node[zone] (z24)  at (3.0,-1.25) {$z_{24}$};
	\node[zone] (z35)  at (4.5,-1.25) {$z_{35}$};
	\node[zone] (z456) at (6.0,-1.25) {$z_{456}$};
	\node[zone] (z6)   at (7.5,-1.25) {$z_{6}$};
	
	\node[unit] (u1)   at (7.5,-3.5)  {$u_{1}$};
	\node[unit] (u2)   at (6.0,-3.5)  {$u_{2}$};
	\node[unit] (u3)   at (4.5,-3.5)  {$u_{3}$};
	\node[unit] (u4)   at (3.0,-3.5)  {$u_{4}$};
	\node (u5)         at (1.5,-3.5) {\small$\ucap=2$};
	\node (u6)         at (0,-3.5)   {\small$\iucap=2$};
	
	\draw 
	(z1)   -- (s1)
	(z35)  -- (s3)
	(z35)  -- (s5)
	(z456) -- (s4)
	(z456) -- (s5)
	(z456) -- (s6)
	(z123) -- (s1)
	(z123) -- (s2)
	(z123) -- (s3)
	(z6)   -- (s6)
	(z24)  -- (s2)
	(z24)  -- (s4);
	\end{tikzpicture}
\end{minipage}
\hfill
\begin{minipage}[b]{0.3\textwidth}
	\begin{tikzpicture}[x=1.7em,y=2.5em,baseline=0pt,rotate=90]
	\tikzstyle{sensor} = [draw, circle, inner sep=0em, minimum width=1.5em]
	\tikzstyle{zone} =   [draw, rectangle, inner sep=.2em, minimum width=2em, rounded corners=0.2em]
	\tikzstyle{unit} =   [draw, rectangle, inner sep=.2em, minimum width=2em]
	
	\node[sensor] (s1) at (0.0,1.25) {$s_1$};
	\node[sensor] (s2) at (1.5,1.25) {$s_2$};
	\node[sensor] (s3) at (3.0,1.25) {$s_3$};
	\node[sensor] (s4) at (4.5,1.25) {$s_4$};
	\node[sensor] (s5) at (6.0,1.25) {$s_5$};
	\node[sensor] (s6) at (7.5,1.25) {$s_6$};

	\node[zone] (z1)   at (0.0,-1.25) {$z_{1}$};
	\node[zone] (z123) at (1.5,-1.25) {$z_{123}$};
	\node[zone] (z24)  at (3.0,-1.25) {$z_{24}$};
	\node[zone] (z35)  at (4.5,-1.25) {$z_{35}$};
	\node[zone] (z456)   at (6.0,-1.25) {$z_{456}$};
	\node[zone] (z6)  at (7.5,-1.25) {$z_{6}$};
	
	\node[unit] (u1) at (1.50, 0) {$u_1$};
	\node[unit] (u2) at (3.75, 0) {$u_2$};
	\node[unit] (u3) at (6.00, 0) {$u_3$};

	\draw 
	(z1)    -- (u1) -- (s1)
	(z123)  -- (u1) -- (s2)
	(z24)   -- (u2) -- (s3)
	(z35)   -- (u2) -- (s4)
	(z6)    -- (u3) -- (s5)
	(z456)  -- (u3) -- (s6);
	
	\draw (u1) -- (u2) -- (u3);
	\end{tikzpicture}
\end{minipage}
\caption{Sample security monitoring system layout with the row length $n=2$ (right), a derived PUP instance (center), and one of its solutions (right).} \label{fig:pup_example}    
\end{figure}
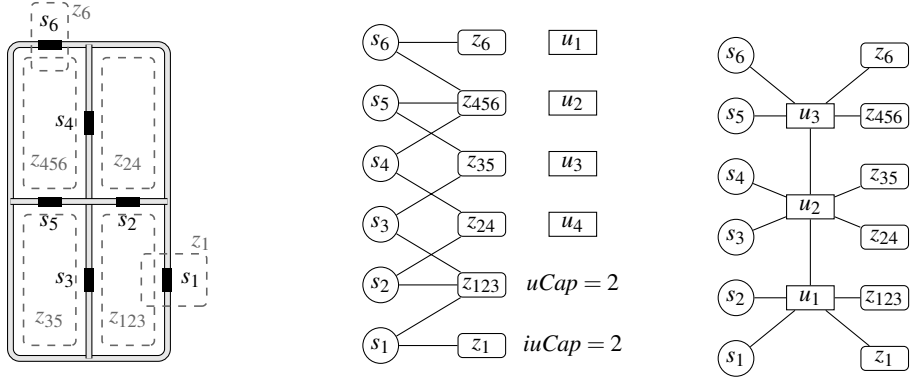

\end{example}

Formally, the partner unit problem can be defined as follows.

\begin{definition}[PUP problem]
Let $P=\langle Z, S, E, U, \ucap, \iucap \rangle$, 
where $Z$ and $S$ are sets of zones resp.\ sensors, 
$E\subseteq Z\times S$ is a set of zone-to-sensors relations, and 
$U$ is a set of units with $\ucap$ many zone/sensor ports and $\iucap$
many inter-unit ports.
A solution is a graph $L=\langle Z\cup S\cup U, H \rangle$
with edges $H\subseteq (Z\times U) \cup (S\times U) \cup (U\times U)$
representing zone-to-unit, sensor-to-unit, and unit-to-unit relations
such that
\begin{enumerate}
	\item Each zone and sensor is connected to exactly one unit;
	\item Each unit is connected to at most $\ucap$ zones/sensors
          resp.\ $\iucap$ units (called partner units);
  \item If a zone $z$ and a sensor $s$ are connected to different units,
  i.e., $(z,u), (s,u') \in H$ where $u\neq u'$, then $(z,s)\in E$
  implies $(u,u') \in H$.
\end{enumerate}
\end{definition}

\begin{example}[PUP security application, cont.]
  A possible solution graph for our example PUP problem instance is shown in \reffig{fig:pup_example}. This solution uses three units that form a simple network allowing for the fast communication between door sensors and related security zones. 
\end{example}

\myParagraph{PUP in stream reasoning applications.}
Various security and safety applications of CPSs can be represented as PUP. 
Stream reasoning in these applications is used to monitor and
(re)configure a CPS e.g.\ in case of administration actions, failures of system components, etc. 
For instance, an administrator may temporarily change a configuration of security zones for some event, a door sensor may fail, or a security zone can be deactivated for building maintenance. 
In such situations, changes in a CPS are continuously communicated to a stream reasoner, which
should find a new configuration of the CPS for its new state. 
Solutions of new instances must be communicated back to the CPS as fast as possible to ensure the best results. 

\myParagraph{Instance generation.}
To simulate a stream of events from a CPS, we implemented a generator that applies random modifications to a given PUP instance, representing a CPS for a security monitoring application. 
The instance is selected from the family of \emph{double} PUP instances representing security monitoring systems \cite{DBLP:conf/cpaior/AschingerDFGJRT11}.
It consists of two rows of rooms arranged on a grid and connected by doors wherever two rooms meet as in \reffig{fig:pup_example}. 
The sensors are installed at each door and the security zones are laid over the rooms.
We measure the instance size by a row length $n$, i.e., there are $2n$ zones and $3n-2$ sensors. 
In our evaluation all experiments were performed on instances with row lengths $n \in \{6,\ldots,11\}$.

The instances are modified by applying \emph{mutation operators} that represent events registered by a monitoring component of a CPS stream reasoning system. 
Such monitoring can easily be done by LARS-based 
solutions as shown e.g.\ in \cite{DBLP:conf/icc/BeckBDEHS17,DBLP:journals/tplp/EiterOS19}. 
All such events are encoded by a set of ground atoms representing modifications to the current PUP instance.

To model real-world scenarios appearing in a security CPS, we use the following mutations:
\begin{enumerate*}[label=(\emph{m}\arabic*)]
  \item disabling a random zone, 
  \item disabling a random sensor,
  \item restoring the original problem.
\end{enumerate*} 
Mutations $m1$ and $m2$ correspond to rooms 
out of order resp.\ doors becoming blocked and represent faults in the system,
while mutation $m3$ corresponds to restoring the initial CPS state.

The generator uses randomization to simulate the randomly occurring
events in the CPS. 
For the mutations $m1$ and $m2$, we assume that the involved zones and
sensors are selected according to Zipf's law, which is often observed
in practice when some components of a CPS are more likely to fail than
the others (cf.\ Example \ref{ex:fault_distr}). The mutation $m3$ is activated according to a Bernoulli distribution with a probability $p$. 

To ensure the repeatability of each experiment for different
algorithms and thus the comparability of results, the generator first creates a random ordering $e_1,\ldots,e_n$ of the zones/sensors and builds a probability distribution such that the frequency of $e_i$ is inverse proportional to the rank $i$.
That is, the probability to select $e_i$ is proportional to $1/i^{\alpha}$,
where the parameter $\alpha$ controls the skew of the distribution towards 
$e_1,\ldots,e_{i-1}$. 
\revision{
For each experiment, we selected two values $\alpha \in \{2.2, 0.7\}$.
These values ensure that $x$\% of the elements are
picked $(100 - x)$\% of the time, for $x \in \{20, 40\}$.
}
For instance, for $\alpha=2.2$ the sampling follows the 
Pareto Principle, 
by which 20\% of the components fail in 80\% of all cases. In general, the smaller the value of $\alpha$ the closer is Zipf's law to the uniform distribution.
Note that the selected $\alpha$ values were specifically determined for our experiments to approximate the target selection ratios as close as possible.

\myParagraph{Experiment.}
The goal of the experiment was to evaluate the performance of
Alg.~\ref{algo:management} while tackling various situations occurring
in a CPS as well as
the impact of the caching strategies. 
We generated a set of stream-reasoning instances using the mutation schema
$[m1,m3,m2,m3]$ where $m3$ was applied with $p=0.8$, i.e., there is an $\approx$80\% chance that all modifications by previous mutations will be repaired.
These settings resulted in PUP instances with $\approx$3 modifications. 
Each CPS simulation has a sequence of 256 PUP instances generated by repeating the mutation schema. 

\begin{figure}[bt]
  \centering
  \includegraphics[width=\linewidth]{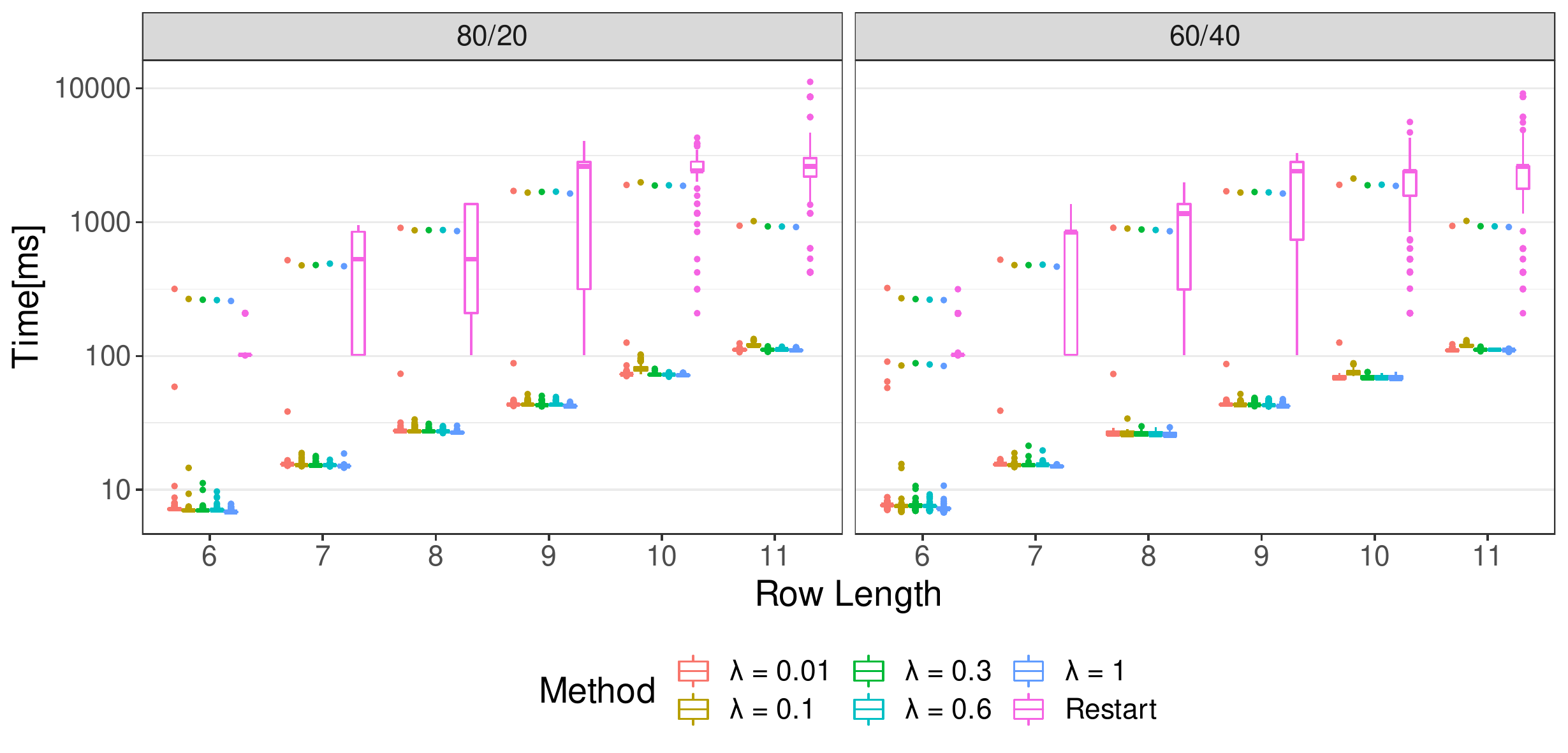}
  \caption{\label{fig:single-fault-boxplot} Results of the experiment
    with the partner unit problem (PUP)}
\end{figure}

Since the optimal learning rate $\lambda$ is unknown, we conducted a
number of experiments with different $\lambda$-values, which results are shown in \reffig{fig:single-fault-boxplot}. 
The \mrestart method has the worst performance, while the RL method
needs for all values of $\lambda$ comparable time to find a solution for each incoming PUP instance.
As it turned out during the experiment, a relatively small subset of all learned constraints (mostly binary or ternary) had a dramatic impact on the performance of the solver. All learning strategies could identify those constraints and always unfreeze them in the solver.
Therefore, in all RL experiments the reasoning time has a very small variance as the span of the 1\textsuperscript{st} and the 3\textsuperscript{rd} quartiles, indicated by boxes, as well as of the min and the max values, shown by whiskers, is very small. The topmost outlier, represented by a point, corresponds to the solving time of the first instance, which comprises the overgrounding time. 
However, the overgrounding is executed only once when the stream reasoning system is starting up and thus has no influence on the reasoning performance during operation time.
The other outliers for smaller learning rates $\lambda \leq 0.1$ occurred because the learner was too conservative and could not react fast enough to data changes in the stream.
This behavior can be explained by the selected exploration strategy. The learners with $\lambda > 0.1$ were able to update the optimistic initial weights of learned constraint to good estimates of expected rewards faster than the learners with $\lambda \leq 0.1$.
Further experiments indicate that RL can solve streaming PUP instances up to row length 20 with the same median time as \mrestart for row length 11. 
Moreover, \mrestart spent most of the time for model search and not 
for program grounding, as shown in Table \ref{tab:grounding}.

\begin{table}[b]
  \centering
  \begin{tabular}{rcccccc}\hline
    Row length & 6     &     7  &     8  &     9  &    10  &    11  \\ \hline
    \mrestart &  9.99  &  12.73  &  16.02  &  20.50  &  25.73  &  32.62  \\
    RL with $\lambda=0.1$ &  28.04   &  47.40   &  77.13   &  122.80   &  187.55   &  266.35   \\ \hline
    \end{tabular}%
  \caption{Mean grounding times measured in milliseconds for the PUP instances \label{tab:grounding}} 
\end{table}

\begin{figure}[bt]
  \centering
  \includegraphics[width=\linewidth]{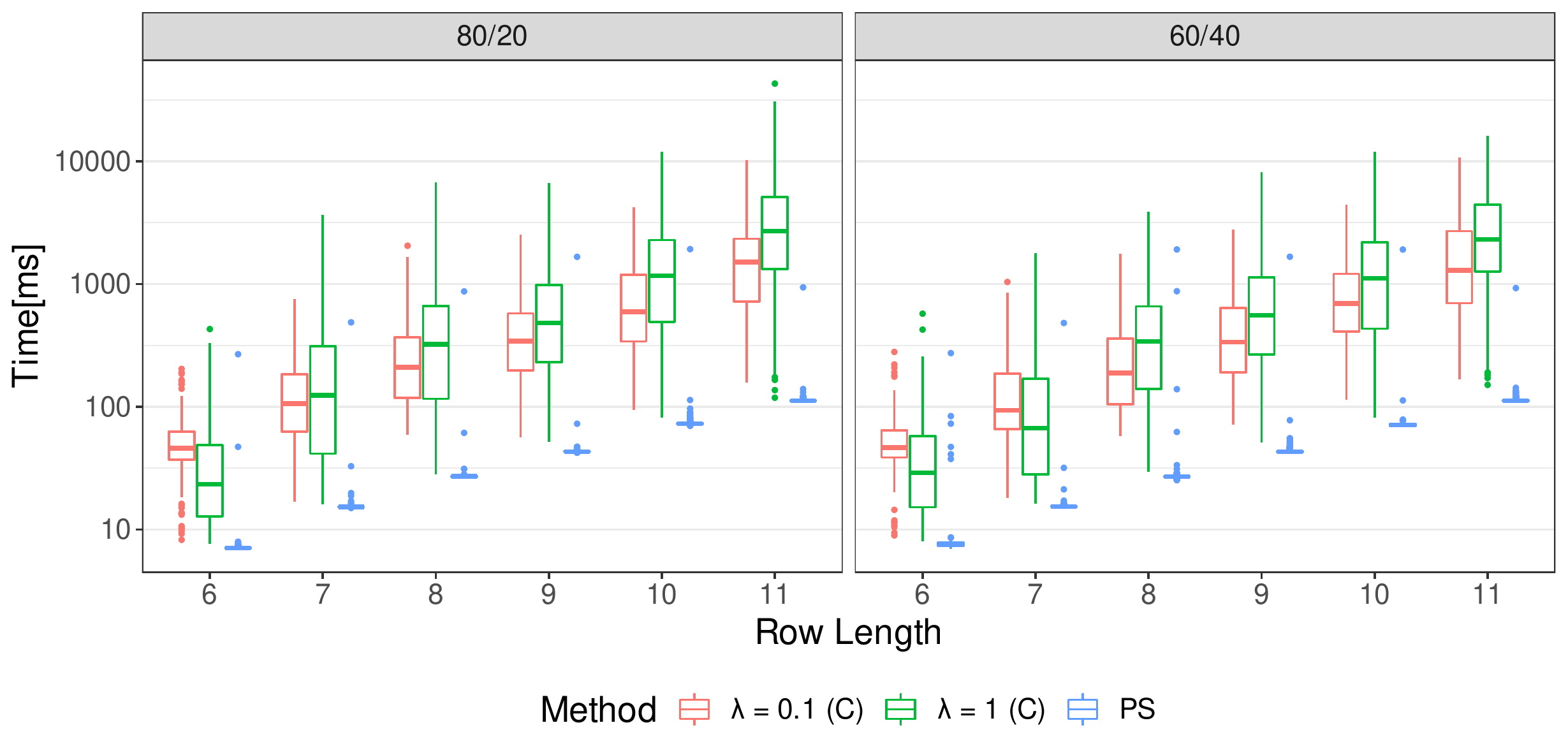}
  \caption{\label{fig:single-fault-mods} Impact of the individual
    caching strategies in the PUP experiment}
\end{figure}

Furthermore, we performed experiments isolating each caching strategy to measure its impact on the performance of the stream reasoner.
In this experiment, RL used only constraints managed according to 
their strategies, labeled with (C) in \reffig{fig:single-fault-mods}, whereas PS keeps all constraints frozen thus forcing the solver to rely only on progress saving. 
The most interesting results were obtained for RL with $\lambda=1$ and $\lambda=0.1$. They indicate that progress saving has the largest impact on the reasoner performance.
As consecutive instances are in this scenario very similar,
only small parts of a legacy configuration become incoherent wrt.\ incoming facts.
Progress saving (PS) allows the solver to rapidly reconstruct coherent parts of the model and then focus solely on the repair of its small incoherent part.
However, the increased number of outliers indicates that in some cases repairs were not easy to find compared to RL (see \reffig{fig:single-fault-boxplot}).
Among the ``constraint-only'' strategies $\lambda=0.1$ appears to be a better learning rate on most of the instances.
The performance of the learner initialized with $\lambda=1$ degrades in line with the skewness of the
distribution (cases 80/20 and 60/40) that allows for more different
changes in an incoming instance.
As this learner prefers constraints relevant to the model for the initial system state, it cannot select proper constraints if a rare event occurs.

We also made a similar experiment with $\alpha = 3.64$ where 10\% of the components fail in 90\% of all cases, and an experiment with single modifications, by setting $p=1$ in the generator.
The results are quite similar to those for multiple modifications (see
online appendix).\footnote{\revision{\url{http://distributed-stream-reasoner.ainf.at}}}

\subsection{n-Queens Completion}

The n-Queens Completion problem (QC) is well-known to be NP-complete
\cite{DBLP:journals/jair/GentJN17} and
an interesting benchmark for stream reasoning systems \cite{DBLP:journals/tplp/EiterOS19},
defined as follows: \emph{given} an $n\times n$ chessboard and a set
$Q=\{q_1,\ldots,q_k\}$, $k<n$, of queens $q_i$ placed on it,
such that no $q_i\neq q_j \in Q$ attack each other, 
\emph{place} all remaining $n-|Q|$ queens on the board 
with that property.

\myParagraph{Instance generation.}
The instance generator was implemented in a similar way as for PUP. 
First, 
it creates an initial QC instance by generating a set $Q$ of
$\left\lfloor 0.4 \cdot n \right\rfloor$ many queens 
which are randomly 
positioned on the board
according to Corollary 15 in \cite{DBLP:journals/jair/GentJN17}, which guarantees the generation of a satisfiable QC instance.
The instance is modified using the following mutations:
\begin{enumerate*}[label=(\emph{m}\arabic*)]
  \item rotate a board counterclockwise at $90^{\circ}$; 
  \item place a random non-attacking queen on the board; and
  \item restore the original problem.
\end{enumerate*}
The column in which a queen is added by $m2$ is selected wrt.\ Zipf's
law, where our experiment used $\alpha=1.35$. 
This value corresponds to a rather moderate skewness of the underlying distribution that leads to a more uniform selection of a column where a queen is placed.
A row for a new queen was computed according to Corollary 15 where valid placements were evaluated in random order.
Finally, the value of the restart probability for $m3$ was defined as $p=0.95$.
The generator was then used to create streaming test instances 
by iteratively applying the mutation schema $[m1,m2,m1,m3]$.
Note that according to \citeN{DBLP:journals/jair/GentJN17}, this generator is not a sustainable one, i.e., that can be used to find really difficult instances for modern solvers.
Nevertheless, given the performance requirements to stream reasoners, the resulting instances are suitable for our experiments.

\begin{figure}[t]
  \centering
  \includegraphics[width=\linewidth]{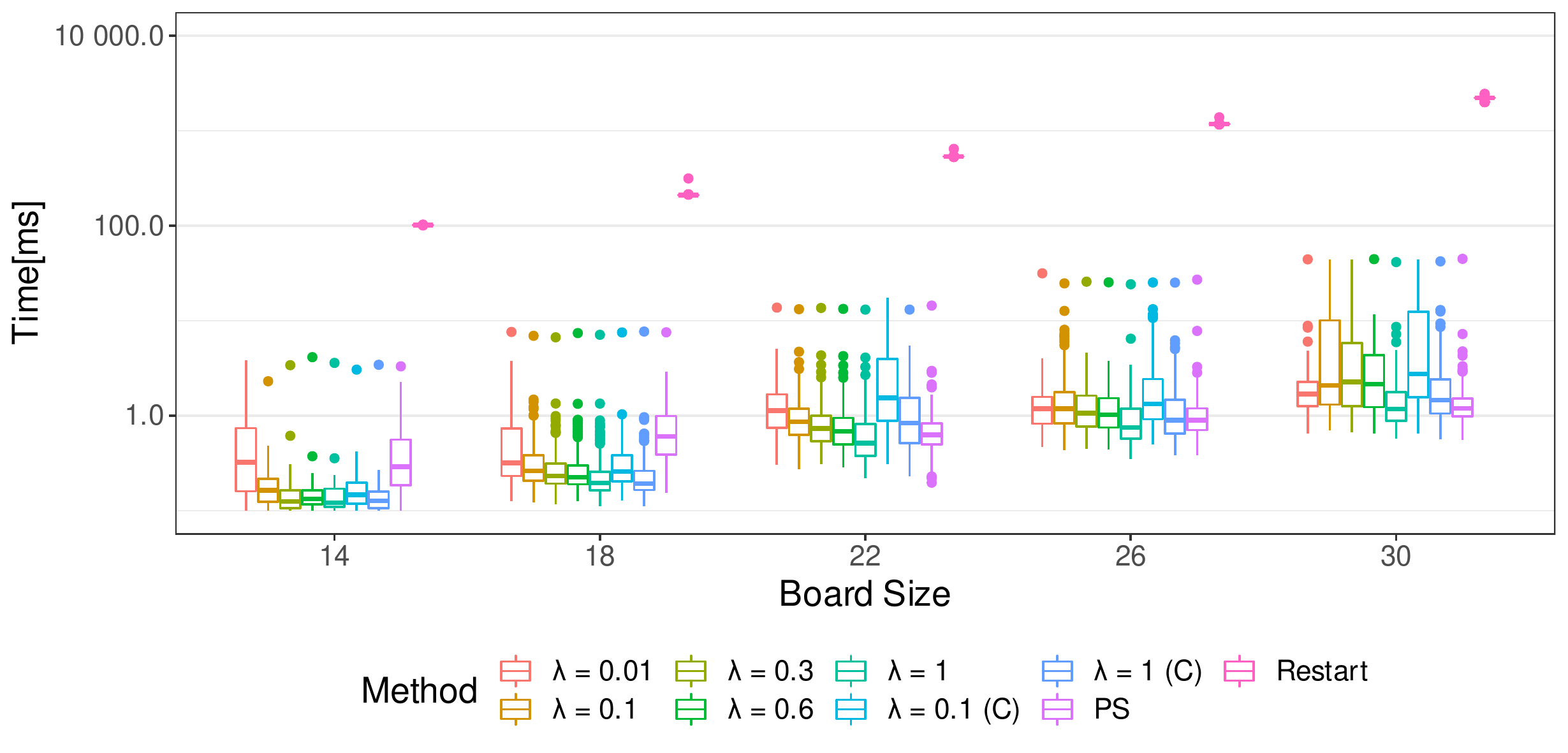}
  \caption{\label{fig:nqueens-boxplot} Results of the experiment with the n-Queens Completion problem}
\end{figure}

\myParagraph{Experiment.}
\revision{
We generated four streaming instances for $n \in \{14,18,22,26,30\}$
with 256 QC instances each and evaluated them for the same learning rates as in the previous section. 
The results, shown in \reffig{fig:nqueens-boxplot}, are that as in the previous experiment the performance of RL with different learning rates is comparable for the same reason 
and significantly outperforms \mrestart:
in the largest experiment, the worst RL result of $\approx 80$ ms is more than 120 times better than the solving time of \mrestart.
Also, we can observe that for the large instances $n > 18$ the performance of learners that is comparable for the most optimistic $\lambda = 1$ and the most conservative $\lambda=0.01$ learning rates. 
The remaining learning rates prevented the learner to stay focused on the most useful constraints.
However, the experiment also showed that the impact of individual caching strategies depends on the initial placement of queens.
Thus, for the first two instances, the modifications introduced by $m1$ were large and the progress saving method (PS in \reffig{fig:nqueens-boxplot}) was unable to help the solver to reconstruct the solution.
The learned clauses reused by RL without progress saving, labeled with (C) in \reffig{fig:nqueens-boxplot}, provided more information to the propagation algorithm for smaller instances with $n \leq 18$.
Nevertheless, PS turned out to be useful while solving the
larger instances with $n\geq 22$.
}
In general, the experiment shows that it is quite hard to
predict which learning rate would be more useful during the solving process.
An extensive study of applying different learning strategies for
balancing of various caching strategies remains for future work. 

\section{Conclusion}

In this paper, we have discussed caching techniques used in modern ASP solvers and presented two approaches to the management of learned constraints based on reinforcement learning. The evaluation results that we presented indicate that proper reuse of data obtained while solving one instance from a data stream can significantly improve the performance of modern solvers while solving subsequent instances, and hence of ASP-based stream reasoning engines on top of them.

Moreover, the experiments provided support for the findings of previous research on the application of truth maintenance system techniques in stream reasoners like \ticker \cite{DBLP:conf/icc/BeckBDEHS17}.  Progress saving -- a JTMS-like caching strategy -- appears to be very useful in monitoring applications of technical systems.  Since massive changes or failures are rarely observed in such environments under normal conditions, data delivered by such systems in subsequent time points is highly interrelated.  Therefore, a solver using progress saving can easily restore consistent parts of a previous model and focus only on repairing of a rather small number of remaining unsatisfiable assignments.  However, current progress saving methods are less versatile in comparison to JTMS techniques when the number of possible constants in a program is large.  In such situations, they usually cannot select which literals must be removed from the cache as it grows in size.  This finding opens an interesting direction of future research, especially in conjunction with predictive overgrounding techniques \cite{DBLP:journals/tplp/CalimeriIPPZ19}.

The positive effect of caching of learned constraints was observed in situations when data in the input stream caused many inconsistent assignments in the existing model.
In such situations learned constraints were able to provide valuable 
information to the solver that could be fruitfully used to reduce its search time. However, our findings also showed that the application of reinforcement learning in this area must be studied in more detail.
In our future work, we are going to focus on automated identification of reward functions that work best for a particular encoding in the LARS language and we shall consider experiments with highly dynamic domains, such as cooperative intelligent transport systems.

\subsection*{Acknowledgments}
The authors would like to thank the anonymous reviewers for their helpful comments that helped us to improve the paper a lot.
This work was conducted in the scope of the research project 
\textit{DynaCon (FFG-PNr.: 861263)} funded by the Austrian Federal Ministry 
of Transport, Innovation and Technology (BMVIT) under the program ``ICT of the Future" between 2017 and 2020 (see \url{https://iktderzukunft.at/en/} for more information).


\begin{thebibliography}{}

  \bibitem[\protect\citeauthoryear{Adams}{Adams}{1984}]{DBLP:journals/ibmrd/Adams84}
  {\sc Adams, E.~N.} 1984.
  \newblock Optimizing preventive service of software products.
  \newblock {\em {IBM} J. Res. Dev.\/}~{\em 28,\/}~1, 2--14.
  
  \bibitem[\protect\citeauthoryear{Alviano, Dodaro, Faber, Leone, and
    Ricca}{Alviano et~al\mbox{.}}{2013}]{DBLP:conf/lpnmr/AlvianoDFLR13}
  {\sc Alviano, M.}, {\sc Dodaro, C.}, {\sc Faber, W.}, {\sc Leone, N.}, {\sc
    and} {\sc Ricca, F.} 2013.
  \newblock {WASP:} {A} native {ASP} solver based on constraint learning.
  \newblock In {\em {LPNMR}}. 54--66.
  
  \bibitem[\protect\citeauthoryear{Alviano, Dodaro, Leone, and Ricca}{Alviano
    et~al\mbox{.}}{2015}]{DBLP:conf/lpnmr/AlvianoDLR15}
  {\sc Alviano, M.}, {\sc Dodaro, C.}, {\sc Leone, N.}, {\sc and} {\sc Ricca, F.}
    2015.
  \newblock Advances in {WASP}.
  \newblock In {\em {LPNMR}}. 40--54.
  
  \bibitem[\protect\citeauthoryear{Anantharam, Varaiya, and Walrand}{Anantharam
    et~al\mbox{.}}{1987}]{anantharamAsymptoticallyEfficientAllocation1987}
  {\sc Anantharam, V.}, {\sc Varaiya, P.}, {\sc and} {\sc Walrand, J.} 1987.
  \newblock Asymptotically efficient allocation rules for the multiarmed bandit
    problem with multiple plays-{Part} {I}: {I}.{I}.{D}. rewards.
  \newblock {\em IEEE Trans. on Automatic Control\/}~{\em 32,\/}~11, 968--976.
  
  \bibitem[\protect\citeauthoryear{Aschinger, Drescher, Friedrich, Gottlob,
    Jeavons, Ryabokon, and Thorstensen}{Aschinger
    et~al\mbox{.}}{2011}]{DBLP:conf/cpaior/AschingerDFGJRT11}
  {\sc Aschinger, M.}, {\sc Drescher, C.}, {\sc Friedrich, G.}, {\sc Gottlob,
    G.}, {\sc Jeavons, P.}, {\sc Ryabokon, A.}, {\sc and} {\sc Thorstensen, E.}
    2011.
  \newblock Optimization methods for the partner units problem.
  \newblock In {\em {CPAIOR}}. 4--19.
  
  \bibitem[\protect\citeauthoryear{Audemard and Simon}{Audemard and
    Simon}{2009}]{DBLP:conf/ijcai/AudemardS09}
  {\sc Audemard, G.} {\sc and} {\sc Simon, L.} 2009.
  \newblock Predicting learnt clauses quality in modern {SAT} solvers.
  \newblock In {\em {IJCAI}}. 399--404.
  
  \bibitem[\protect\citeauthoryear{Audemard and Simon}{Audemard and
    Simon}{2018}]{DBLP:journals/ijait/AudemardS18}
  {\sc Audemard, G.} {\sc and} {\sc Simon, L.} 2018.
  \newblock On the glucose {SAT} solver.
  \newblock {\em Int. J. Artif. Intell. Tools\/}~{\em 27,\/}~1, 1--25.
  
  \bibitem[\protect\citeauthoryear{Bazoobandi, Beck, and Urbani}{Bazoobandi
    et~al\mbox{.}}{2017}]{DBLP:conf/semweb/BazoobandiBU17}
  {\sc Bazoobandi, H.~R.}, {\sc Beck, H.}, {\sc and} {\sc Urbani, J.} 2017.
  \newblock Expressive stream reasoning with laser.
  \newblock In {\em ISWC}. 87--103.
  
  \bibitem[\protect\citeauthoryear{Beck, Bierbaumer, Dao{-}Tran, Eiter,
    Hellwagner, and Schekotihin}{Beck
    et~al\mbox{.}}{2017}]{DBLP:conf/icc/BeckBDEHS17}
  {\sc Beck, H.}, {\sc Bierbaumer, B.}, {\sc Dao{-}Tran, M.}, {\sc Eiter, T.},
    {\sc Hellwagner, H.}, {\sc and} {\sc Schekotihin, K.} 2017.
  \newblock Stream reasoning-based control of caching strategies in {CCN}
    routers.
  \newblock In {\em {ICC}}. {IEEE}, 1--6.
  
  \bibitem[\protect\citeauthoryear{Beck, Dao{-}Tran, and Eiter}{Beck
    et~al\mbox{.}}{2015}]{DBLP:conf/ijcai/BeckDE15}
  {\sc Beck, H.}, {\sc Dao{-}Tran, M.}, {\sc and} {\sc Eiter, T.} 2015.
  \newblock Answer update for rule-based stream reasoning.
  \newblock In {\em {IJCAI}}. {AAAI} Press, 2741--2747.
  
  \bibitem[\protect\citeauthoryear{Beck, Dao{-}Tran, and Eiter}{Beck
    et~al\mbox{.}}{2018}]{DBLP:journals/ai/BeckDE18}
  {\sc Beck, H.}, {\sc Dao{-}Tran, M.}, {\sc and} {\sc Eiter, T.} 2018.
  \newblock {LARS:} {A} logic-based framework for analytic reasoning over
    streams.
  \newblock {\em Artif. Intell.\/}~{\em 261}, 16--70.
  
  \bibitem[\protect\citeauthoryear{Beck, Eiter, and Folie}{Beck
    et~al\mbox{.}}{2017}]{DBLP:journals/tplp/BeckEB17}
  {\sc Beck, H.}, {\sc Eiter, T.}, {\sc and} {\sc Folie, C.} 2017.
  \newblock Ticker: {A} system for incremental asp-based stream reasoning.
  \newblock {\em {TPLP}\/}~{\em 17,\/}~5-6, 744--763.
  
  \bibitem[\protect\citeauthoryear{Calimeri, Ianni, Pacenza, Perri, and
    Zangari}{Calimeri et~al\mbox{.}}{2019}]{DBLP:journals/tplp/CalimeriIPPZ19}
  {\sc Calimeri, F.}, {\sc Ianni, G.}, {\sc Pacenza, F.}, {\sc Perri, S.}, {\sc
    and} {\sc Zangari, J.} 2019.
  \newblock Incremental answer set programming with overgrounding.
  \newblock {\em Theory Pract. Log. Program.\/}~{\em 19,\/}~5-6, 957--973.
  
  \bibitem[\protect\citeauthoryear{de~Kleer}{de~Kleer}{1986}]{DBLP:journals/ai/Kleer86}
  {\sc de~Kleer, J.} 1986.
  \newblock An assumption-based {TMS}.
  \newblock {\em Artif. Intell.\/}~{\em 28,\/}~2, 127--162.
  
  \bibitem[\protect\citeauthoryear{Doyle}{Doyle}{1979}]{DBLP:journals/ai/Doyle79}
  {\sc Doyle, J.} 1979.
  \newblock A truth maintenance system.
  \newblock {\em Artif. Intell.\/}~{\em 12,\/}~3, 231--272.
  
  \bibitem[\protect\citeauthoryear{Eiter, Ogris, and Schekotihin}{Eiter
    et~al\mbox{.}}{2019}]{DBLP:journals/tplp/EiterOS19}
  {\sc Eiter, T.}, {\sc Ogris, P.}, {\sc and} {\sc Schekotihin, K.} 2019.
  \newblock A distributed approach to {LARS} stream reasoning (system paper).
  \newblock {\em Theory Pract. Log. Program.\/}~{\em 19,\/}~5-6, 974--989.
  
  \bibitem[\protect\citeauthoryear{Gai, Krishnamachari, and Jain}{Gai
    et~al\mbox{.}}{2012}]{gaiCombinatorialNetworkOptimization2012a}
  {\sc Gai, Y.}, {\sc Krishnamachari, B.}, {\sc and} {\sc Jain, R.} 2012.
  \newblock Combinatorial {Network} {Optimization} {With} {Unknown} {Variables}:
    {Multi}-{Armed} {Bandits} {With} {Linear} {Rewards} and {Individual}
    {Observations}.
  \newblock {\em IEEE/ACM Transactions on Networking\/}~{\em 20,\/}~5,
    1466--1478.
  
  \bibitem[\protect\citeauthoryear{Gaschnig}{Gaschnig}{1979}]{gaschnigjohnPerformanceMeasurementAnalysis1979}
  {\sc Gaschnig, J.} 1979.
  \newblock Performance measurement and analysis of certain search algorithms.
  \newblock Ph.D. thesis, Carnegie Mellon University, Pittsburgh, PA, USA.
  
  \bibitem[\protect\citeauthoryear{Gebser, Grote, Kaminski, Obermeier, Sabuncu,
    and Schaub}{Gebser et~al\mbox{.}}{2012}]{DBLP:conf/kr/GebserGKOSS12}
  {\sc Gebser, M.}, {\sc Grote, T.}, {\sc Kaminski, R.}, {\sc Obermeier, P.},
    {\sc Sabuncu, O.}, {\sc and} {\sc Schaub, T.} 2012.
  \newblock Stream reasoning with answer set programming: Preliminary report.
  \newblock In {\em {KR}}. {AAAI} Press.
  
  \bibitem[\protect\citeauthoryear{Gebser, Kaminski, Kaufmann, and Schaub}{Gebser
    et~al\mbox{.}}{2019}]{DBLP:journals/tplp/GebserKKS19}
  {\sc Gebser, M.}, {\sc Kaminski, R.}, {\sc Kaufmann, B.}, {\sc and} {\sc
    Schaub, T.} 2019.
  \newblock Multi-shot {ASP} solving with clingo.
  \newblock {\em Theory Pract. Log. Program.\/}~{\em 19,\/}~1, 27--82.
  
  \bibitem[\protect\citeauthoryear{Gelfond and Lifschitz}{Gelfond and
    Lifschitz}{1988}]{DBLP:conf/iclp/GelfondL88}
  {\sc Gelfond, M.} {\sc and} {\sc Lifschitz, V.} 1988.
  \newblock The stable model semantics for logic programming.
  \newblock In {\em {ICLP/SLP}}. {MIT} Press, 1070--1080.
  
  \bibitem[\protect\citeauthoryear{Gent, Jefferson, and Nightingale}{Gent
    et~al\mbox{.}}{2017}]{DBLP:journals/jair/GentJN17}
  {\sc Gent, I.~P.}, {\sc Jefferson, C.}, {\sc and} {\sc Nightingale, P.} 2017.
  \newblock Complexity of n-queens completion.
  \newblock {\em J. Artif. Intell. Res.\/}~{\em 59}, 815--848.
  
  \bibitem[\protect\citeauthoryear{Gomes, Selman, and Kautz}{Gomes
    et~al\mbox{.}}{1998}]{DBLP:conf/aaai/GomesSK98}
  {\sc Gomes, C.~P.}, {\sc Selman, B.}, {\sc and} {\sc Kautz, H.~A.} 1998.
  \newblock Boosting combinatorial search through randomization.
  \newblock In {\em {AAAI/IAAI}}. {AAAI} Press / The {MIT} Press, 431--437.
  
  \bibitem[\protect\citeauthoryear{Hehenberger, Vogel{-}Heuser, Bradley, Eynard,
    Tomiyama, and Achiche}{Hehenberger
    et~al\mbox{.}}{2016}]{DBLP:journals/cii/HehenbergerVBET16}
  {\sc Hehenberger, P.}, {\sc Vogel{-}Heuser, B.}, {\sc Bradley, D.}, {\sc
    Eynard, B.}, {\sc Tomiyama, T.}, {\sc and} {\sc Achiche, S.} 2016.
  \newblock Design, modelling, simulation and integration of cyber physical
    systems: Methods and applications.
  \newblock {\em Comput. Ind.\/}~{\em 82}, 273--289.
  
  \bibitem[\protect\citeauthoryear{Huang}{Huang}{2007}]{DBLP:conf/ijcai/Huang07}
  {\sc Huang, J.} 2007.
  \newblock The effect of restarts on the efficiency of clause learning.
  \newblock In {\em {IJCAI}}. 2318--2323.
  
  \bibitem[\protect\citeauthoryear{Kaufmann, Leone, Perri, and Schaub}{Kaufmann
    et~al\mbox{.}}{2016}]{DBLP:journals/aim/KaufmannLPS16}
  {\sc Kaufmann, B.}, {\sc Leone, N.}, {\sc Perri, S.}, {\sc and} {\sc Schaub,
    T.} 2016.
  \newblock Grounding and solving in answer set programming.
  \newblock {\em {AI} Magazine\/}~{\em 37,\/}~3, 25--32.
  
  \bibitem[\protect\citeauthoryear{Nadel and Ryvchin}{Nadel and
    Ryvchin}{2012}]{DBLP:conf/sat/NadelR12}
  {\sc Nadel, A.} {\sc and} {\sc Ryvchin, V.} 2012.
  \newblock Efficient {SAT} solving under assumptions.
  \newblock In {\em {SAT}}. 242--255.
  
  \bibitem[\protect\citeauthoryear{Pipatsrisawat and Darwiche}{Pipatsrisawat and
    Darwiche}{2007}]{DBLP:conf/sat/PipatsrisawatD07}
  {\sc Pipatsrisawat, K.} {\sc and} {\sc Darwiche, A.} 2007.
  \newblock A lightweight component caching scheme for satisfiability solvers.
  \newblock In {\em {SAT}}. 294--299.
  
  \bibitem[\protect\citeauthoryear{{Ratasich}, {Khalid}, {Geissler}, {Grosu},
    {Shafique}, and {Bartocci}}{{Ratasich} et~al\mbox{.}}{2019}]{8606923}
  {\sc {Ratasich}, D.}, {\sc {Khalid}, F.}, {\sc {Geissler}, F.}, {\sc {Grosu},
    R.}, {\sc {Shafique}, M.}, {\sc and} {\sc {Bartocci}, E.} 2019.
  \newblock A roadmap toward the resilient internet of things for cyber-physical
    systems.
  \newblock {\em IEEE Access\/}~{\em 7}, 13260--13283.
  
  \bibitem[\protect\citeauthoryear{Rossi and Rossini}{Rossi and
    Rossini}{2012}]{DBLP:conf/infocom/RossiR12}
  {\sc Rossi, D.} {\sc and} {\sc Rossini, G.} 2012.
  \newblock On sizing {CCN} content stores by exploiting topological information.
  \newblock In {\em {INFOCOM} Workshops}. {IEEE}, 280--285.
  
  \bibitem[\protect\citeauthoryear{Silva and Sakallah}{Silva and
    Sakallah}{1996}]{DBLP:conf/ictai/SilvaS96}
  {\sc Silva, J. P.~M.} {\sc and} {\sc Sakallah, K.~A.} 1996.
  \newblock Conflict analysis in search algorithms for satisfiability.
  \newblock In {\em {ICTAI}}. {IEEE} Computer Society, 467--469.
  
  \bibitem[\protect\citeauthoryear{Sutton}{Sutton}{1995}]{DBLP:conf/nips/Sutton95}
  {\sc Sutton, R.~S.} 1995.
  \newblock Generalization in reinforcement learning: Successful examples using
    sparse coarse coding.
  \newblock In {\em {NIPS}}. {MIT} Press, 1038--1044.
  
  \bibitem[\protect\citeauthoryear{Sutton and Barto}{Sutton and
    Barto}{2018}]{Sutton2018}
  {\sc Sutton, R.~S.} {\sc and} {\sc Barto, A.~G.} 2018.
  \newblock {\em Reinforcement Learning: An Introduction\/}, 2nd ed.
  
  \bibitem[\protect\citeauthoryear{Swift and Warren}{Swift and
    Warren}{2012}]{DBLP:journals/tplp/SwiftW12}
  {\sc Swift, T.} {\sc and} {\sc Warren, D.~S.} 2012.
  \newblock {XSB:} extending prolog with tabled logic programming.
  \newblock {\em Theory Pract. Log. Program.\/}~{\em 12,\/}~1-2, 157--187.
  
  \end{thebibliography}

\label{lastpage}
\end{document}